\documentclass[10pt,twocolumn,letterpaper]{article}
\pdfoutput=1
\usepackage{cvpr}
\usepackage[pagebackref=true,breaklinks=true,letterpaper=true,colorlinks,bookmarks=false]{hyperref}
\usepackage{times}
\usepackage{epsfig}
\usepackage{graphicx}
\usepackage{amsmath}
\usepackage{amssymb}
\usepackage{bbm}
\usepackage{url}
\DeclareMathOperator*{\argmax}{argmax} 
\usepackage{comment}
\usepackage{diagbox}
\usepackage{colortbl}
\usepackage{amsthm}
\usepackage{amsfonts}
\usepackage{verbatim}
\usepackage{multirow}
\graphicspath{ {./} }

\cvprfinalcopy 

\def\cvprPaperID{6219} 

\newcommand{\thetav}{\boldsymbol\theta}

\ifcvprfinal\fi
\begin{document}

\title{Auto-Annotation Quality Prediction for Semi-Supervised Learning with Ensembles}

\author{Dror Simon \qquad Miriam Farber\qquad Roman Goldenberg\\
Amazon.com\\
{\tt\small drorsimo/mffarber/romang@amazon.com}
}

\maketitle

\begin{abstract}
   Auto-annotation by ensemble of models is an efficient method of learning on unlabeled data. Wrong or inaccurate annotations generated by the ensemble may lead to performance degradation of the trained model. To deal with this problem we propose filtering the auto-labeled data using a trained model that predicts the quality of the annotation from the degree of consensus between ensemble models. Using semantic segmentation as an example, we show the advantage of the proposed auto-annotation filtering over training on data contaminated with inaccurate labels. 
   
   Moreover, our experimental results show that in the case of semantic segmentation, the performance of a state-of-the-art model can be achieved by training it with only a fraction (30$\%$) of the original manually labeled data set, and replacing the rest with the auto-annotated, quality filtered labels. 
\end{abstract}

\section{Introduction}
Semi-supervised learning, using the combination of a smaller set of labeled data and a larger set of unlabeled data, is becoming increasingly important with the growing capacity of trained models and their tasks complexity. Higher capacity models require more training data, and complex tasks make manual annotation more labor intensive. Ensembles of models have been successfully used for automatic annotation of unlabeled data \cite{CaruanaBucilaCN06, Zhou04}. In self-training scenarios, multiple instances of the target model trained on the labeled data form an ensemble. The ensemble, which is said to be more accurate than a single model, then labels the unlabeled data. In model-compression scenarios, ensembles extract the knowledge from a set of more powerful pre-trained models.

Using ensembles for unlabeled data annotation, however, raises one issue that is often not properly addressed: the annotation quality. In general, the ensemble cannot guarantee that the annotations it generates meet the quality bar. Using wrong or inaccurate annotations for training may negatively affect the target model. 

In this work we propose a method for predicting the quality of the annotations generated by an ensemble. The approach uses a model  trained to assess the quality of the generated annotation from the degree of consensus between ensemble models. 

We propose to refine the auto-labeled data set by discarding samples with low predicted annotation quality. We show that training on a refined reduced set is advantageous over using a lager set, contaminated with inaccurate data.

The main contributions of the paper are:
\begin{itemize}
    \item We introduce an automatic filtering of auto-annotations generated by ensembles, using a trained model that predicts annotation quality from ensemble models degree of consensus.
    \item We propose an auto-annotation quality control scheme for semantic segmentation, which filters bad labels at pixel level, yielding a refined partial image labeling.
    \item We demonstrate that training semantic segmentation model with quality filtering is advantageous over training with no filtering.
    \item We show how the proposed method achieves the same accuracy as a state-of-the-art model, while manually annotating of only a fraction (30$\%$) of the training set.
\end{itemize}

\section{Related work}
Ever since is was demonstrated that ensembles of models can boost the accuracy \cite{Hansen90}, and the reasons ensembles outperform any single model they are composed of were identified \cite{dietterich2000ensemble}, they have been used extensively to achieve state-of-the-art performance. A large body of work exists on the subject, reviewed in \cite{valentini2002ensembles,re2011ensemble, jurek2014survey}.  

According to the taxonomy developed in \cite{valentini2002ensembles, kuncheva2004combining}, the ensemble approaches are divided into non-generative - using a fixed set of pre-designed models, and generative - generating models by acting on the base learning algorithm or on the structure of the training set. While the approach we propose can be generally applied to both types, in the experiments presented here we used ensemble generation by bagging \cite{breiman1996bagging} the same model trained with different permutations and augmentations of the training set. Other techniques include bagging by varying parameter settings \cite{Caruana2004}, a low computational cost snapshot ensembles \cite{laine2016temporal, GaoHuang2017, chen2017checkpoint} and many others.

Here we combine bagging with a complimentary approach to accuracy boosting - ensembling by input transformations. Variants of it were used in various computer vision tasks including object detection \cite{viola2001rapid,dollar2014fast}, image classification \cite{krizhevsky2012imagenet}, recognition \cite{sermanet2013overfeat} and keypoints detection \cite{Radosavovic}.

Using ensembles of models for self-training was proposed in \cite{CaruanaBucilaCN06}, where a knowledge distillation is performed on unlabeled data by an ensemble trained on a smaller set of labeled data. The idea of self-labeling goes back to 1965 \cite{scudder1965probability}, and since then was a subject of research in semi-supervised learning \cite{zhu2006semi}. The benefits of using ensembles for semi-supervised learning are advocated in \cite{zhou2009semi}.

Here we propose a model distillation regime where the quality of the labeling generated by an ensemble is estimated by an additional, second-level model. This is closely related to the stacked generalization  \cite{wolpert1992stacked} meta-learning technique \cite{duin2000experiments}. Unlike the method in \cite{HintonGeoffreyVinyalsOriolDean2015} that uses soft labels for knowledge distillation, we train a network to filter out bad and unreliable labels. 

Decontaminating noisy labels from training data was explored in \cite{barandela2000decontamination, brodley1999identifying}. Those approaches assume no prior knowledge about the training set labels, whereas in our case the labels are generated for initially unlabeled data by an ensemble. In this work we leverage the degree of consensus between ensemble models as additional information to assess labels quality.

We demonstrate the effectiveness of the proposed technique on the example of semantic segmentation \cite{garcia2017review}. Prior work on not fully supervised semantic segmentation include weakly and partially supervised techniques \cite{khoreva2017simple, papandreou2015weakly, vezhnevets2010towards}, self-supervision \cite{zhan2017mix} and ensemble knowledge transfer \cite{nigam2018ensemble}. To the best of our knowledge, this is the first work that performs semi-supervised training of semantic segmentation model on auto-annotated unlabeled data, generated by ensembles with quality filtering.

\section{Annotation quality prediction}
\label{sec:quality_prediction}
Let $f: \mathbb{X} \rightarrow \mathbb{Y}$ be the target model to be trained. We use an ensemble of models $e = (f_1,\dots,f_k), f_j: \mathbb{X} \rightarrow \mathbb{Y}$, to automatically label the unlabeled data for training the target model $f$. Models in $e$ are trained on a labeled data set $\mathbb{S}= \{(x^{(i)},\bar{y}^{(i)})\}, x^{(i)}\in\mathbb{X}, \bar{y}^{(i)}\in\mathbb{Y}$. 

We use the ensemble to generate labels for the large unlabeled set $\mathbb{U}= \{x^{(i)}\}$ in the following way:
\begin{itemize}
\item Run $e: \mathbb{X}\rightarrow\mathbb{Y}^k$ on $\mathbb{U}$ to generate vectors of labels $\mathbb{L}=\{(y^{(i)}_1,\dots,y^{(i)}_k)\}$.
\item Apply a fusing function $g: \mathbb{Y}^k\rightarrow\mathbb{Y}$ to combine the ensemble labels into a single label: $\widehat{\mathbb{L}} = \{\hat{y}^{(i)}\}= g(\mathbb{L}) = (g \circ e)(\mathbb{U})$ 
\end{itemize}
The $g$ function can be implemented by a plethora of ensemble fusion methods \cite{re2011ensemble}. It can either generate a novel labeling $\hat{y}$ from ensemble outputs $(y_1,\dots,y_k)$ , or select one of them, i.e. $\hat{y} = g(y_1,\dots,y_k) = y_j, j\in\{1,2,\ldots,k\}$.

We can then train the target model on the generated labeled set $\mathbb{T} = \{\mathbb{U},\widehat{\mathbb{L}}\} = \{(x^{(i)},\hat{y}^{(i)})\}$. In a general case, some of the automatically generated labels in $\widehat{\mathbb{L}}$ are expected to be wrong. Therefore,  a supposedly better approach would be to remove the corresponding data samples from $\mathbb{T}$.

To filter out bad labels, one can use an ad-hoc method. For instance, if ensemble models reports a confidence level associated with the output, it can be used to filter out the low quality annotations. Such filtering is quite simple in some scenarios, e.g. for object/image classification tasks, while in others, e.g. object detection, it is becoming less trivial. The confidence associated with each detected object does not provide any information on the likelihood of missing an object.  For example, to deal with this issue in keypoint and object detections, \cite{Radosavovic} filter out false positives by thresholding the confidence score to get “the average number of annotated instances per unlabeled image” roughly equal to “the average number of instances per labeled image”. 

In a more principled way, we propose to train a function $q$, that predicts the quality of the labeling generated by an ensemble, based on the degree of consensus between ensemble models. Function $q:\mathbb{Y}^k\rightarrow \{0,1\} $ receives the ensemble output $(y_1,\dots,y_k)$ and generates a quality score in $\{0,1\}$. This is similar to Wolpert's ensemble stacking approach \cite{wolpert1992stacked}, but instead of merging ensemble outputs to generate a fused labeling, the $q$ function predicts the labeling quality for a fixed fusor $g$. The function $q$ is trained using a labeled data set $\mathbb{Q} = \{(x^{(i)},\bar{y}^{(i)})\}$.
For a data sample $(x,\bar{y})\in \mathbb{Q}$, the input for the $q$ model is the ensemble output $e(x)$ and the ground truth is the indicator function $\mathbbm{1}_{{(g\circ e)(x)=\bar{y}}}$.






We then use $q$ to filter the auto-annotated set $\mathbb{T}$ by discarding data samples with low predicted  annotation quality to yield a refined labeled data set $\mathbb{T}^{*} = \{(x,\hat{y})\in \mathbb{T} \mid q(e(x)) = 1\}$

\section{Training with auto-annotations and the EM algorithm}
\label{sec:EM}
In this section, we step back to perform a more principled analysis of the auto-annotation procedure, providing a mathematical justification to the benefit of using auto-annotated data from an optimization process point of view. We extend our approach by viewing the auto-annotation procedure as an iterative process, in which the following steps are repeated several times. 
\begin{enumerate}
  \item  Start from a manually labeled data set $\mathbb{S}$.
  \item  Train an ensemble of models on the available data set.
  \item  Using the ensemble, obtain auto-annotations on an unlabeled data set.
  \item  Apply quality filtering to remove poorly annotated data, thus improving the quality of the auto-annotations.
  \item  Add the auto-annotations to the available data set.
  \item  Repeat steps 2-5 several times to improve model's performance, until performance is stabilized.
\end{enumerate}
In this paper, we consider a special case of this process, analyzing the model's improvement after one iteration of the steps above. To make the discussion in this section more concrete, we concentrate on a classification problem. 

We perform our analysis under a semi-supervised framework. This is similar, in concept, to the analysis in \cite{amini2002semi}. In this scheme, using the notations of the previous section, we are given a set of labeled samples $\mathbb{S}= \{(x^{(i)},\bar{y}^{(i)})\}$ and a set of unlabeled samples $\mathbb{U}= \{x^{(t)}\}$. Our goal is to fit a model $p_{\thetav}(x,y)$ to the data, characterized by a set of parameters $\thetav$. In order to do that, we would like to optimize the log likelihood function $\ell(\thetav)$, defined via
\begin{equation}
\sum_{(x^{(i)},\bar{y}^{(i)})\in\mathbb{S}}\log p \left(x^{(i)},\bar{y}^{(i)};\thetav\right)+\sum_{x^{(t)} \in\mathbb{U}}\log p \left(x^{(t)};\thetav\right).
\end{equation}
If all our data was labeled, we could have approached this optimization problem using methods such as gradient descent. However, part of our data is unlabeled, resulting in two types of parameters we need to fit: The parameters $\thetav$ and the missing labels for the data in $\mathbb{U}$. Since in this discussion we focus on a classification problem, we can view these missing labels as class probability vectors, associated with each $x \in \mathbb{U}$. Thus, we would like to find both $\thetav$ and the mentioned probability vectors. 

In order to do that, we can utilize a well-known likelihood maximization method: Expectation Maximization (EM) algorithm \cite{maxlikeem}, which is an iterative approach that converges to a local maximum. We first obtain an initial estimation for $\thetav$, by fitting the model (in our case, training the network) using only the labeled data. Afterwards, each iteration of EM consists of two steps:
\begin{itemize}
\item In the E-step, the posterior probabilities of each class $1 \leq c \leq C$ (where $C$ is the total number of classes) are estimated for each sample $x^{(t)} \in \mathbb{U}$, using the model and its parameters. The posterior probability vector is of the form
\begin{equation}
    \bigg(p\left(\hat{y}^{(t)}=1\middle|x^{(t)};\thetav\right),\ldots,p\left(\hat{y}^{(t)}=C\middle|x^{(t)};\thetav\right)\bigg).
\end{equation}
This step is performed by activating the model on a sample and inferring the probabilities of each class.
\item In the M-step, we optimize a lower bound of the likelihood function w.r.t. the parameters of the model (this is a well known lower bound. See, for example, section 2 in \cite{maxlikeem_lowerbound}). Thus, $\thetav$ equals to $\argmax$ (over $\thetav$) of

\begin{equation}
\sum_{x^{(m)}}\sum_{c=1}^C p\left(y=c\middle|x^{(m)};\thetav\right) \log p \left(x^{(m)},y=c;\thetav\right).
\end{equation}
In the summation above, $x^{(m)}$ is summed over all the elements in $\mathbb{U}$ and the first coordinate of the elements in $\mathbb{S}$. For the latter case, $y$ is chosen to be $\bar{y}^{(m)}$. For the former case, $y$ enumerates over all possible classes.

This step corresponds to training the model on the inferred posterior probabilities. In the case where the model is a neural network, applying the maximization step is equivalent to training the network, using the cross-entropy loss.
\end{itemize}

In our method, we add an additional intermediate classification phase (C-step) to the EM algorithm, which we refer to as \emph{auto-annotation}. After the expectation phase, a per-class probability exists for each sample. We leverage these probabilities by modifying the posteriors via taking the most probable class, i.e. $p\left(y=c\middle|x^{(m)};\thetav\right)=1$ if $c$ is the class with the largest posterior probability, and otherwise it equals 0. 

To conclude, we first train a model on the labeled samples, leading to an initial set of parameters $\thetav$. Then, we apply the first two steps of our revised EM algorithm. We apply the trained model on unlabeled samples (E-step), and then we classify them to produce auto-annotations for the unlabeled data (C-step). Finally, the M-step consists of retraining the model, this time using labeled samples along with newly auto-annotated ones. Note that, while we use the auto-annotated labels for the unlabeled samples, we keep the original annotations for the labeled ones.

Clearly, a satisfying convergence of the described approach relies on the accuracy of the auto-annotations themselves. When the auto-annotations are too noisy, the model is not expected to improve its generalization  \cite{noisylabels}.

To reduce the noise caused by false auto-annotations, and improve the convergence, we propose to use two means. The first is to strengthen the model itself by averaging an ensemble of models, as suggested in \cite{schapire1995strenght, naftaly1997optimal}. Furthermore, as described in the previous section, we train an additional classifier $q$ aimed to predict false auto-annotations and eliminate them from the retraining process. 

\section{Semi-supervised semantic segmentation with auto-annotation quality prediction } 

To demonstrate the proposed approach, we implement a semi-supervised training of a semantic segmentation model using auto-annotation with ensemble of models and quality filtering. We chose the task of semantic segmentation since creating manual annotations for this task is extremely labor intensive, resulting in a relatively limited amount of such annotations. Therefore, highly accurate auto-annotations could be especially useful for this task.

To populate the ensemble we use both multiple models and data augmentation, following the data distillation method presented in \cite{Radosavovic}. First, we train the same model multiple (three) times using different parameters initialization and training samples reshuffling. In addition, each model is fed six augmented versions of the input image: two horizontal flips $\times$ three scales (x0.5, x1.0, and x1.5) . This effectively corresponds to an ensemble of size 18.

In the experiments described in the next section we merge ensemble results into a single label using a simple softmax averaging \cite{schapire1995strenght, van1997combining, kuncheva1997application, naftaly1997optimal}. We refer to a collection of such labels as unfiltered auto annotations. The fusing function $g$ is defined as
\begin{equation}
\label{eq:softmax_average} 
g(\sigma_1,\dots,\sigma_k) = \argmax_{i\in [1,\dots,C]} \sum_{j=1}^{k} \sigma_j,
\end{equation}
where $C$ is the number of classes in the semantic segmentation model, and $\sigma_j \in \mathbb{R}^C$ is a softmax class probability vector generated by the $j$-th model of an ensemble of size $k$.

Interestingly, since semantic segmentation models can be trained on a partially labeled image, we do not necessarily need to accept or discard the image labeling as a whole. Instead, we can do it selectively, by making a decision per pixel. Pixels with unreliable labeling are marked as a special "ignore" class in the labels mask and do not contribute to the gradient back-propagation during the training.

In addition, if the segmentation model generates not only the class labels, but also class probabilities, we can use those as inputs to the quality filter $q$. Here we implement $q$ using a convolutional neural network (CNN) that receives $k\times C $ input channels - $(\sigma_1,\dots,\sigma_k)$ and outputs a quality mask in $\{0,1\}$.

The network has four hidden conv-ReLU layers. The first layer has 40 output channels while the rest of them have 20. This is followed by the output conv-sigmoid layer with one output channel.

\begin{figure*}
\begin{center}
\includegraphics[width=0.3\textwidth]{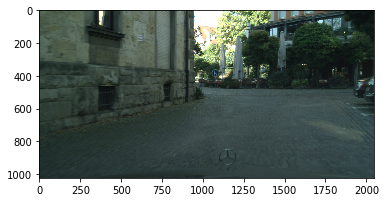}
\includegraphics[width=0.3\textwidth]{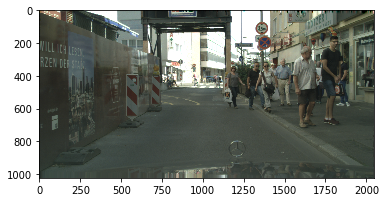}
\includegraphics[width=0.3\textwidth]{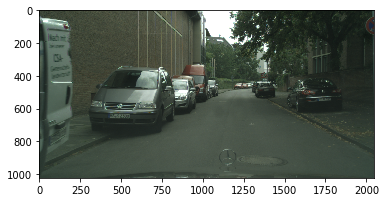}
\includegraphics[width=0.3\textwidth]{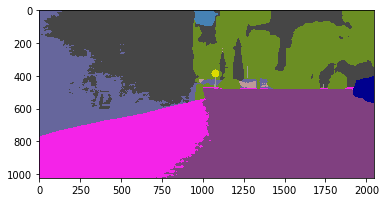}
\includegraphics[width=0.3\textwidth]{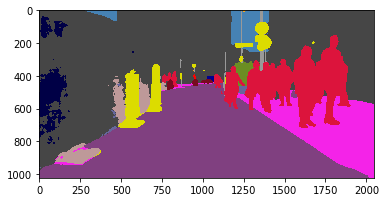}
\includegraphics[width=0.3\textwidth]{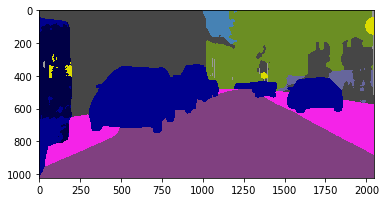}
\includegraphics[width=0.3\textwidth]{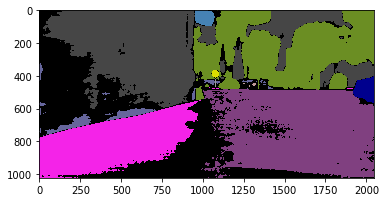}
\includegraphics[width=0.3\textwidth]{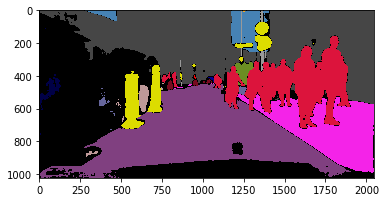}
\includegraphics[width=0.3\textwidth]{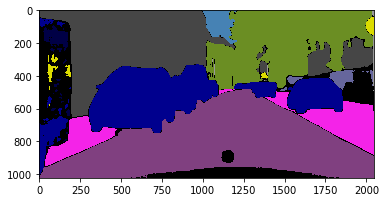}
\includegraphics[width=0.3\textwidth]{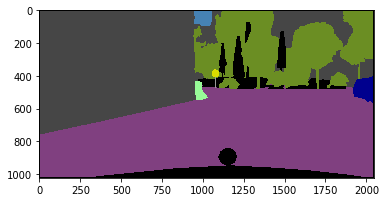}
\includegraphics[width=0.3\textwidth]{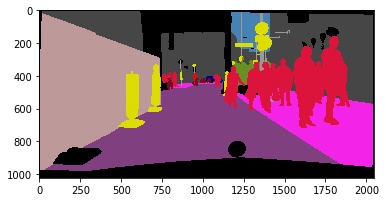}
\includegraphics[width=0.3\textwidth]{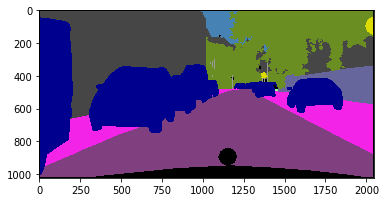}
\end{center}
   \caption{Rows - top to bottom: (1) Original image, (2) unfiltered auto-annotations, (3) quality filtered auto-annotations, (4) ground truth annotation. Columns: three different examples. Black pixels in the fourth row represent void classes that are not counted as part of the 19 classes and do not contribute to the \textit{mIoU} score. Black pixels in the third row represent pixels that are masked out by the quality filter.}
\label{fig:examples}
\end{figure*}
\section{Experiments}
In this section we demonstrate the strength of the proposed quality prediction approach. We show that a model trained on a combination of manually labeled data and quality filtered auto-annotations achieves better performance than a model trained on a combination of manually labeled data and unfiltered auto annotations. In fact, we show that quality filtering allows replacing significant amount of manually-annotated images by auto-annotated images, without any degradation in accuracy.
\subsection{Training procedure}
The experiments are performed with DeepLab network \footnote{https://github.com/tensorflow/models/tree/master/research/deeplab \label{ftn:deeplab}} \cite{deeplabv3plus2018}, a state of the art network for semantic segmentation. The training protocols involve 90000 iterations, batch size of 12, learning rate of 0.01, output stride of 16, and decoder output stride of 4. We follow the same cropping convention as recommended in \cite{deeplabv3}, and use crop size of 769 during training. Finally, we use Xception65 \cite{Xception} pretrained on ImageNet \cite{ImageNet} as a backbone. The evaluation results, both on the validation and the test sets are obtained using single scale and output stride of 16.
\subsection{Data set}
\label{subsec:datasplit}
We evaluate our approach on the pixel-level semantic labeling task of Cityscapes data set \cite{cityscapes}. This data set has 19 semantic labels (and additional void labels that are not used for evaluation), and consists of  5000  images, which are split into training, validation and test sets of sizes 2975, 500, and 1525 respectively. We  report \textit{mIoU} (mean \textit{IOU}) scores and also \textit{IoU} scores of each one of the 19 classes following the Cityscapes definitions. Results are reported on the validation and the test sets.

Our experiments involve several splits of data. In all the splits below, all the 19 classes are represented.
\begin{itemize}
\item \textbf{Cityscapes full training set} - 2975 labeled images that form the Cityscapes training data.
\item \textbf{Cityscapes extra set} - 3600 images sampled from the 20000 coarsely annotated Cityscapes data set. We do not use any available labels from this data set, only the images themselves.
\item \textbf{Cityscapes small training set} - 30$\%$ of the labeled images in the Cityscapes training set. 
\item \textbf{Cityscapes tiny training set} - 15$\%$ of the labeled images in the Cityscapes training set. 
\end{itemize}
As the 19 classes are unbalanced by nature, when choosing subsets of data sets (training or unlabeled), it is important to ensure that smaller classes are well represented. To this end we utilize the following automatic approach: For the labeled data set, we choose subsets (30$\%$ and 15$\%$, in our experiments) based on automatic filtering of the manual labels (choosing only images in which the number of labeled pixels of certain classes is high enough). For the unlabeled set, we first generate auto-annotations via ensemble of models, and then apply the same selection logic as for the labeled set, using auto-annotations instead of manual labels.
\subsection{Training with and without quality prediction}
\label{subsec:training_with_without_quality_pred}
In this experiment we show that a model trained on manually labeled data + quality filtered auto-annotations outperforms a model  trained on manually labeled data + unfiltered auto annotations. The former model surpasses the latter by 0.8$\%$ in \textit{mIoU}.

The experiment, whose results are summarized in table~\ref{tab:results_on_qual_pred}, consists of the following steps: We first train the network on the Cityscapes tiny training set 3 times, and evaluate the network's performance (the best accuracy among the 3 is indicated in the first column in the table). We then produce unfiltered auto-annotations for the Cityscapes extra set by fusing ensemble results following equation (\ref{eq:softmax_average}). The ensemble consists of the 3 trained models together with 6 augmentations associated with each model, effectively corresponds to an ensemble of size 18 as described in the previous section. 

We proceed by performing two additional trainings: first, we train the network on a data set which consists of the Cityscapes tiny training set and the unfiltered auto-annotations described above. The network's performance is indicated in the second column of the table. Next, we refine the auto-annotations by applying the quality filter, described in section~\ref{sec:quality_prediction}. The quality filter is trained on the Cityscapes tiny training set. Finally, we train the network on a data set which consists of the Cityscapes tiny training set and the quality filtered auto-annotations. The network's performance is indicated in the third column of the table. 

\begin{table}
    \begin{tabular}{| l | l | l |}
    \hline
    \shortstack{Manual \\ annotations \\ only } & \shortstack{Manual +\\ unfiltered \\ auto-annotations} & \shortstack{Manual + quality \\ filtered \\ auto-annotations}  \\ \hline
     74.0$\%$& 75.5$\%$ & 76.3$\%$ \\
    \hline
    \end{tabular}
    \caption{\textit{mIoU} on the validation set of the experiment depicted in section~\ref{subsec:training_with_without_quality_pred}.}
    \label{tab:results_on_qual_pred}
\end{table}

As shown in the table, adding unfiltered auto-annotated data improves the trained model accuracy by 1.5$\%$. Adding quality filtering improves the performance by additional 0.8$\%$, which leads to overall improvement of 2.3$\%$.

We chose the tiny training set for this experiment, as in this scenario the auto-annotated data forms large portion of the training data. In fact, the auto-annotated data forms roughly 90$\%$ of the training data used for the training procedures depicted in the second and third columns of table~\ref{tab:results_on_qual_pred}. This mimics the real world scenario in which available unlabeled data sets are much larger than their manually-labeled counterparts. Our approach shows that such data sets can be utilized effectively to improve network's performance.   

\subsection{Training with and without quality prediction -- per class examination}
The experiment from the previous section shows a clear improvement in performance when filtered auto-annotated data is added to the training data. In this section, we perform a class-wise IoU examination, obtained via the experiment above. Based on the results presented in table~\ref{tab:classwise_iou_performance_val}, we arrive to the following conclusions:
\begin{enumerate}
  \item  A model trained on manually labeled data + quality filtered auto-annotations outperforms a model trained on manually labeled data + unfiltered auto annotations on each one of the 19 classes.
  \item A model trained on manually labeled data + quality filtered auto-annotations performs significantly better on underrepresented classes, compared to a model trained on manually labeled data only, adding up to $\sim$20$\%$ to the IoU in such classes.    
\end{enumerate}
For the first conclusion, one can see that for each class in table~\ref{tab:classwise_iou_performance_val}, the values in the fifth column surpass the values in the fourth column. This is a strong demonstration of the benefit of the quality filter. The largest gain is demonstrated for the rarest classes (lower rows), with 4$\%$ gain to the most rare class (train). 

The second conclusion is achieved by comparing the third and the fifth columns in table~\ref{tab:classwise_iou_performance_val}. Classes in which the performance of the model trained on manual annotations + quality filtered auto-annotations surpasses the performance of the model trained on manual annotations only are highlighted in green. Those that perform the same are highlighted in yellow. One can see that in 14 out of 19 classes, the performance of the former model was at least as good as the latter. In 11 out of these classes, the performance improved. The most significant improvement is obtained in under-represented classes like train (19.7$\%$ improvement), bus (7.8$\%$ improvement), truck (9.6$\%$ improvement) and fence (4.9$\%$ improvement). One can also see an improvement in the rarest 7 classes.

\begin{table*}[t]
\centering
    \begin{tabular}{| l | l | l | l | l |}
    \hline
    Class & \shortstack{Num of \\ ground truth \\ occurrences} & \shortstack{Manual \\ annotations \\ only } & \shortstack{Manual +\\ unfiltered \\ auto-annotations} & \shortstack{Manual + quality \\ filtered \\ auto-annotations}  \\ \hline
    \rowcolor{yellow} building & 500 & 91.7$\%$ & 91.6$\%$ & 91.7$\%$ \\ \hline
    \rowcolor{green} sidewalk & 499 & 81.8$\%$ & 82.4$\%$ & 83.1$\%$ \\ \hline
    pole & 499 & 63.3$\%$ & 61.2$\%$ & 62.2$\%$ \\ \hline
    \rowcolor{green} road & 498 & 97.7$\%$ & 97.7$\%$ & 97.8$\%$ \\ \hline
    \rowcolor{yellow} vegetation & 493 & 92.1$\%$ & 92.0$\%$ & 92.1$\%$ \\ \hline
    \rowcolor{yellow} traffic sign & 487 & 76.5$\%$ & 76.0$\%$ & 76.5$\%$ \\ \hline
    car & 486 & 94.7$\%$ & 93.7$\%$ & 94.0$\%$ \\ \hline
    \rowcolor{green} sky & 473 & 94.5$\%$ & 94.9$\%$ & 94.9$\%$ \\ \hline
    person & 453 & 81.0$\%$ & 79.3$\%$ & 79.8$\%$ \\ \hline
    \rowcolor{green} fence & 394 & 54.1$\%$ & 57.8$\%$ & 59.0$\%$ \\ \hline
    bicycle & 392 & 75.0$\%$ & 74.0$\%$ & 74.8$\%$ \\ \hline
    traffic light & 385 & 67.6$\%$ & 65.5$\%$ & 65.9$\%$ \\ \hline
    \rowcolor{green} terrain & 351 & 61.4$\%$ & 63.2$\%$ & 63.3$\%$ \\ \hline
    \rowcolor{green} wall & 339 & 49.1$\%$ & 49.6$\%$ & 49.8$\%$ \\ \hline
    \rowcolor{green} rider & 303 & 59.8$\%$ & 59.7$\%$ & 60.3$\%$ \\ \hline
    \rowcolor{green} truck & 187 & 70.6$\%$ & 78.3$\%$ & 80.2$\%$ \\ \hline
    \rowcolor{green} motorcycle & 178 & 61.4$\%$ & 60.8$\%$ & 63.1$\%$ \\ \hline
    \rowcolor{green} bus & 161 & 79.2$\%$ & 85.8$\%$ & 87.0$\%$ \\ \hline
    \rowcolor{green} train & 95 & 55.2$\%$ & 70.9$\%$ & 74.9$\%$ \\ 
    \hline
    \end{tabular}
    \caption{IoU per class on the validation set, as measured in the experiments reported in table~\ref{tab:results_on_qual_pred}. The first column lists the 19 classes in the Cityscapes data set. The second column indicates the number of images in the validation set (out of 500) that contain the specified class. For example, train is the rarest class, available only in 95 of the images, while building is the most common class, available in all the 500 images. Classes are ordered from the most common (first row) to the most rare (last row). Classes in which the performance of the model trained on manual annotations + quality filtered auto-annotations surpasses the performance of the model trained on manual annotations only are highlighted in green. Those that perform the same are highlighted in yellow.}
    \label{tab:classwise_iou_performance_val}
\end{table*}

\subsection{Training with quality prediction on additional data splits}
Following the observed accuracy gain depicted in section~\ref{subsec:training_with_without_quality_pred}, we experiment with  manually-labeled training sets of various sizes. We show that adding quality filtered auto-annotated data improves the model performance in all cases, with the largest improvement achieved for the smallest manually-labeled training set. 

Following the same procedure as in section \ref{subsec:training_with_without_quality_pred}, we train the model first on manually labeled data only, and then on manually labeled data + quality filtered auto-annotated data. We repeat the experiment 3 times by reducing the manually labeled training set from 100$\%$, to 30$\%$ and, finally, to 15$\%$. 


Our results on the validation set are summarized in table \ref{tab:results_on_val}. In table \ref{tab:results_on_test} we also present the accuracy measured on the test set for the models trained on the full manually-labeled training set.  As can be seen in the table, we obtain an improvement of 0.6$\%$ when adding auto-annotated data to the Cityscapes full training set. The improvement increases even further, when the auto-annotated data is added to the small training set and the tiny training set, in which cases we get improvements of 2$\%$ and 2.3$\%$ respectively. 

Moreover, the accuracy of the network trained on the Cityscapes small training set+auto-annotated data is 77$\%$, which is identical to the accuracy of the network trained on Cityscapes full training set. This demonstrates that using our approach, with only 30$\%$ of the available manual annotations, we achieve the same accuracy levels as training with the entire manually annotated training set (saving 70$\%$ of manual annotations).

\begin{table}
    \begin{tabular}{| l | l | l |}
    \hline
    \shortstack{Size of the \\manually labeled\\ training set} & \shortstack{Manually\\ labeled\\ data only} & \shortstack{Manual + quality \\ filtered \\ auto-annotations}  \\ \hline
     Full set (100$\%$) & 77.0$\%$ & 77.6$\%$ \\ \hline
     Small set (30$\%$) & 75.0$\%$ & 77.0$\%$ \\ \hline
     Tiny set (15$\%$) & 74.0$\%$ & 76.3$\%$ \\
    \hline
    \end{tabular}
    \caption{\textit{mIoU}s on the validation set. The data sets in the left column are described in section \ref{subsec:datasplit}. In the middle column, the models were trained on the corresponding manually labeled training set only. In the right column, the models were trained on the corresponding manually labeled set+ quality filtered auto-annotated images from the Cityscapes extra set.}
    \label{tab:results_on_val}
\end{table}

\begin{table}
    \begin{tabular}{| l | l | l |}
    \hline
    \shortstack{Manually\\ labeled\\ training set} & \shortstack{Manually\\ labeled\\ data only} & \shortstack{Manual + quality \\ filtered \\ auto-annotations} \\ \hline
     Full set (100$\%$) & 76.83$\%$ & 77.29$\%$ \\ 
    \hline
    \end{tabular}
    \caption{\textit{mIoU}s on the test set.}
    \label{tab:results_on_test}
\end{table}

\subsection{Performance of the quality filtering}
In this section, we shed some light on the performance of the quality filter itself. In the beginning of the section, we examine retention rate (percentage of pixels that were not masked out by the filter) and auto-annotation precision (both filtered and unfiltered). In the end of the section, we visualize quality filtered auto-annotations. 

Consider the results in table~\ref{tab:quality_filtering}.  The unfiltered and quality filtered auto-annotations of the Cityscapes validation set are used to measure the unfiltered and quality filtered auto-annotation precision respectively. Precision rates increase for all classes when quality filter is applied, showing that our filter indeed learns to identify wrong annotations and to mask them out. The improvement in performance, measured in IoU, of a model trained on manually labeled data + quality filtered auto-annotations versus a model trained on manually labeled data + unfiltered auto annotations is reported in the fifth column. The improvement in IoU is correlated with the improvement in precision (correlation coefficient of 0.48).

Retention rates are indicated in the right-most column in the table. These are the percentage of pixels in non-void locations (based on ground truth) that are retained after applying the quality filter. Overall, 96.2$\%$ of the pixels are retained. These results, in combination with the results in table~\ref{tab:results_on_qual_pred}, show that by applying our quality filter we can improve model's performance while retaining most of the pixels (only 3.8$\%$ of the pixels are masked out).

\begin{table*}[t]
\centering
    \begin{tabular}{| l|l|l|l|l|l|}
    \hline
    \multirow{2}{*}{Class} & \multicolumn{3}{c|}{Annotation precision} & 
    \multirow{2}{*}{\shortstack{Diff between IoU \\ of filtered \\ and unfiltered }} & \multirow{2}{*}{retention}\\
    \cline{2-4}
 &\shortstack{filtered \\ auto-annotations} & \shortstack{unfiltered \\ auto-annotations} & diff \\ \hline
    building & 97.0$\%$ & 95.7$\%$ & 1.3$\%$ & 0.1$\%$  & 97.5$\%$ \\ \hline 
sidewalk & 94.5$\%$ & 89.9$\%$ & 4.7$\%$ & 0.7$\%$  & 86.5$\%$ \\ \hline 
pole & 90.8$\%$ & 81.3$\%$ & 9.6$\%$ & 1.0$\%$  & 70.6$\%$ \\ \hline 
road & 99.7$\%$ & 99.1$\%$ & 0.5$\%$ & 0.1$\%$  & 97.9$\%$ \\ \hline 
vegetation & 96.5$\%$ & 95.4$\%$ & 1.1$\%$ & 0.1$\%$  & 97.5$\%$ \\ \hline 
traffic sign & 95.3$\%$ & 92.4$\%$ & 2.8$\%$ & 0.5$\%$  & 89.5$\%$ \\ \hline 
car & 98.0$\%$ & 97.4$\%$ & 0.6$\%$ & 0.3$\%$  & 97.7$\%$ \\ \hline 
sky & 98.1$\%$ & 97.3$\%$ & 0.8$\%$ & 0.0$\%$  & 98.1$\%$ \\ \hline 
person & 93.2$\%$ & 90.2$\%$ & 3.0$\%$ & 0.5$\%$  & 92.3$\%$ \\ \hline 
fence & 87.6$\%$ & 81.9$\%$ & 5.7$\%$ & 1.2$\%$  & 82.9$\%$ \\ \hline 
bicycle & 91.2$\%$ & 85.4$\%$ & 5.7$\%$ & 0.8$\%$  & 84.7$\%$ \\ \hline 
traffic light & 90.5$\%$ & 86.7$\%$ & 3.8$\%$ & 0.4$\%$  & 86.6$\%$ \\ \hline 
terrain & 88.0$\%$ & 83.8$\%$ & 4.2$\%$ & 0.1$\%$  & 89.0$\%$ \\ \hline 
wall & 85.8$\%$ & 80.4$\%$ & 5.4$\%$ & 0.2$\%$  & 90.4$\%$ \\ \hline 
rider & 84.3$\%$ & 78.2$\%$ & 6.0$\%$ & 0.6$\%$  & 83.7$\%$ \\ \hline 
truck & 96.3$\%$ & 90.2$\%$ & 6.1$\%$ & 1.9$\%$  & 90.8$\%$ \\ \hline 
motorcycle & 93.1$\%$ & 86.3$\%$ & 6.8$\%$ & 2.3$\%$  & 82.2$\%$ \\ \hline 
bus & 92.4$\%$ & 90.4$\%$ & 2.1$\%$ & 1.2$\%$  & 95.8$\%$ \\ \hline 
train & 96.2$\%$ & 90.9$\%$ & 5.3$\%$ & 4.0$\%$  & 86.9$\%$ \\ 
    \hline
    \end{tabular}
    \caption{Auto-annotation precision, improvement in IoU, and retention rates on the validation set. Columns: (1) 19 classes in the Cityscapes data set. (2) Annotation precision of unfiltered auto-annotations on the cityscapes validation set (precision is relative to the ground truth). (3) Annotation precision of quality filtered (according to section~\ref{subsec:training_with_without_quality_pred}) auto-annotations on the validation set. (4) Differences in precision values (filtered minus unfiltered). (5) Model IoU improvement - trained on manually labeled data + quality filtered auto-annotations vs. a model trained on manually labeled data + unfiltered auto annotations - this is the difference between the last two columns from table~\ref{tab:classwise_iou_performance_val}). (6) Filter retention rates - the percentage of pixels in non-void locations (based on ground truth) that are retained after applying the quality filter. Overall, 96.2$\%$ of the pixels are retained.}
    \label{tab:quality_filtering}
\end{table*}

Let us visually inspect unfiltered ensemble annotations and the auto-annotations obtained by applying our quality filter on them. The quality filter has been trained on Cityscapes tiny training set using the models that correspond to the bottom row in table~\ref{tab:results_on_val}. 

Consider the examples in figure~\ref{fig:examples}. The top row depicts three images from the Cityscapes validation set. The corresponding manual labeling (the ground truth) appear in the bottom row. The second row depicts the unfiltered ensemble labeling. Row number three contains the same figures, after undergoing quality filtering. Pixels that are filtered out by the quality filter appear in black. 

One can see that the quality filter identified correctly a substantial amount of errors. For example, a major part of the car's hood is masked out by the filter. The filter also correctly masked out the misclassified parts of the wall in the left and middle examples -- see the purple area (second row), that is correctly filtered out (third row), as can be verified by the ground truth (last row). In the right column example, the misclassified area on the left car is correctly masked out by the quality filter, as well as an area on the right wall. In all three examples, additional more subtle masked out areas can be found in the vicinity of the trees and other small objects. Overall, the filter managed to mask out considerable amount of misclassified pixels, creating more accurate auto-annotated data, which in turn led to an improvement of 0.8$\%$ in the trained model accuracy, as shown in table~\ref{tab:results_on_qual_pred}.

\section{Discussion}

The accuracy gain due to the proposed quality filtering is higher for "less experienced" teacher ensembles: the more mistakes the teacher makes, the more errors the quality filter can fix. There is a trade-off between the requested auto-annotation quality and the amount of generated training data. While in the conducted experiments we used a fixed quality threshold, we would like to explore the influence of the quality-quantity trade-off on the trained model accuracy.

We can further enhance the training process by iterating over the "train the teachers ensemble", "auto-annotate the unlabeled data", and "train the target model" steps, as discussed in section~\ref{sec:EM}. 

Another interesting research direction is using ensemble stacking (instead of softmax averaging) for the fusing function $g$ and building a joint multi-task model for $g$ and $q$ together.
\section{Conclusions}
We propose a generic method for quality prediction of automatic annotations generated by an ensemble of models. We adapt the proposed approach to semantic segmentation by doing label quality filtering at pixel level. We show that refining the auto-annotated training set by discarding data samples with low predicted label quality improves the trained model accuracy. We demonstrate that the performance of the state-of-the-art model can be achieved by training it with only a fraction (30$\%$) of the original manually labeled data set, and replacing the rest with the auto-annotated, quality filtered labels.

{\small
\bibliographystyle{ieee}
\bibliography{egbib}
}

\end{document}